\documentclass[conference]{IEEEtran} 
\usepackage{fix-cm}
\IEEEoverridecommandlockouts
\usepackage{times}
\usepackage{graphicx}
\usepackage{array}
\usepackage{booktabs}
\usepackage{amsmath,amsfonts}
\usepackage{siunitx}
\usepackage{diagbox}

\usepackage[numbers]{natbib}
\usepackage{multicol}
\usepackage[bookmarks=true]{hyperref}

\begin{document}

\title{SoNIC: Safe Social Navigation with Adaptive Conformal Inference and Constrained Reinforcement Learning}
\author{Jianpeng Yao, Xiaopan Zhang$^\dagger$, Yu Xia$^\dagger$, Zejin Wang, Amit K. Roy-Chowdhury, and Jiachen Li*
\thanks{* Corresponding author}
\thanks{$^\dagger$ These authors contributed equally to this paper.}
\thanks{J. Yao, X. Zhang, X. Yu, Z. Wang, A. K. Roy-Chowdhury, and J. Li are with the University of California, Riverside, CA, USA. \{jyao073, xzhang006, yxia072, zejinw, amit.roychowdhury, jiachen.li\}@ucr.edu.}
}

\maketitle

\begin{abstract}
Reinforcement learning (RL) enables social robots to generate trajectories without relying on human-designed rules or interventions, making it generally more effective than rule-based systems in adapting to complex, dynamic real-world scenarios. However, social navigation is a safety-critical task that requires robots to avoid collisions with pedestrians, whereas existing RL-based solutions often fall short of ensuring safety in complex environments.
In this paper, we propose SoNIC, which to the best of our knowledge is the first algorithm that integrates adaptive conformal inference (ACI) with constrained reinforcement learning (CRL) to enable safe policy learning for social navigation.
Specifically, our method not only augments RL observations with ACI-generated nonconformity scores, which inform the agent of the quantified uncertainty but also employs these uncertainty estimates to effectively guide the behaviors of RL agents by using constrained reinforcement learning. This integration regulates the behaviors of RL agents and enables them to handle safety-critical situations. On the standard CrowdNav benchmark, our method achieves a success rate of 96.93\%, which is 11.67\% higher than the previous state-of-the-art RL method and results in 4.5 times fewer collisions and 2.8 times fewer intrusions to ground-truth human future trajectories as well as enhanced robustness in out-of-distribution scenarios. To further validate our approach, we deploy our algorithm on a real robot by developing a ROS2-based navigation system. Our experiments demonstrate that the system can generate robust and socially polite decision-making when interacting with both sparse and dense crowds. The video demos can be found on our project website: \url{https://sonic-social-nav.github.io/}.
\end{abstract}

\IEEEpeerreviewmaketitle

\section{Introduction}
Social robots have been increasingly deployed in a variety of complex real-life environments with dynamic human crowds. 
They are expected to reach their destinations safely, efficiently, and courteously—without being overly aggressive and intruding on humans' future trajectories, nor too cautious to the point of failing their navigation tasks \cite{singamaneni2024survey}.

Reinforcement learning (RL) has shown great potential in this task due to its capability of learning optimal behaviors through interactions with dynamic environments \cite{chen2019crowd,liu2020robot, samsani2021socially,perez2021robot}.
Recent research has shown that RL-based algorithms handle social navigation tasks in complex dynamic environments much better than traditional rule-based methods \cite{liu2023intention}. 

However, since social navigation is a safety-critical task that requires the robots to exhibit extremely safe behaviors to avoid collisions with pedestrians, the performance of current RL solutions is yet to be satisfactory.
Most existing research either presents experimental results in scenarios much simpler than real life \cite{brito2021go} or demonstrates navigation performance in complex scenarios with an unsatisfactory success rate (e.g., below 90\%) or collision rate \cite{liu2023intention}. 
This indicates that current RL-based planners lack the safety and adaptability needed for real-life applications.
\begin{figure}[!tbp]
	\centering
	\includegraphics[width=1.00\columnwidth]{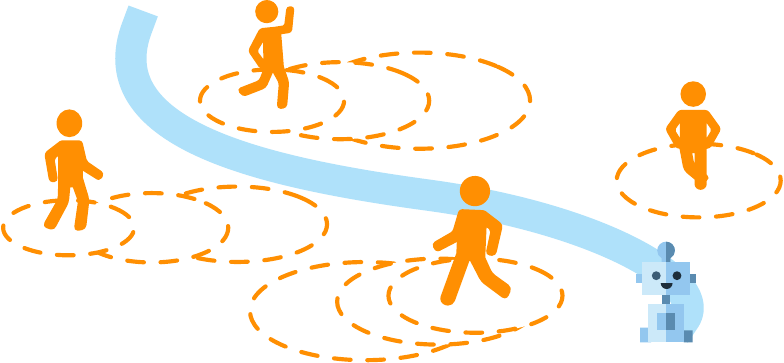}
	\caption{SoNIC employs ACI to generate a spatial buffer around human agents and guide the behaviors of CRL agents to avoid entering the buffer by constraining the cumulative intrusions over each episode.
    }
	\label{fig:teaser}
\end{figure}
Existing methods have two major limitations. 
First, although previous work has employed advanced learning-based prediction models to facilitate the decision making process of robots, these predictions are inherently imperfect. Imperfect predictions may mislead the agents' decisions, thereby degrading the performance of robots in complex scenarios.
We believe that the uncertainties in predictions need to be explicitly quantified and handled \cite{singamaneni2024survey} to mitigate the negative effects of imperfect predictions in an online manner, thereby enabling better adaptation to real-life scenarios.
Second, prior RL work fails to provide an effective way to regulate the behaviors of social robots and prevent them from acting dangerously. Common practices to address this issue include reward shaping and filter-based methods. For reward shaping, researchers incorporate their heuristics and expert knowledge into the reward function for RL policy learning. However, it is challenging to achieve specific behavior patterns purely through reward shaping, as the mapping between rewards and policy behaviors is often intractable \cite{sutton2018reinforcement}, potentially leading to unintended or suboptimal results. 
Regarding filter-based methods, a common approach is to apply safety filters to the output of RL policies, effectively post-editing the actions generated by the RL agents \cite{strawn2023conformal}. However, this combination does not directly enhance the RL learning process, and the insufficient decision making capabilities may still hinder the agents’ performance in complex scenarios.

To address the first limitation, we employ Adaptive Conformal Inference (ACI) \cite{gibbs2021adaptive, gibbs2024conformal} to quantify prediction uncertainty, which provides an area with a pre-defined probability that humans will appear within.
Compared to other conformal methods \cite{angelopoulos2023conformal}, ACI has the advantage of online updating and can adapt to arbitrary distribution shifts, which makes it very suitable for trajectory prediction.
Some works have attempted to combine uncertainty quantification results obtained by other conformal methods with RL by either directly inputting these results into policy networks \cite{huang2023conformal} or forming a filter to process the actions generated by RL \cite{strawn2023conformal}. However, these methods lack the benefits of online updating uncertainties and circumvent the direct guidance on the RL learning process. 
In our work, we not only enhance the observation using ACI-generated nonconformity scores but also directly guide the learning process of the RL agents according to the uncertainty measures.

As for the second limitation, Constrained Reinforcement Learning (CRL) \cite{ray2019benchmarking, ji2023omnisafe} is a promising solution as it can adaptively adjust the weights between reward acquisition and cost avoidance throughout the learning process, and subsequently adjust agent behavior to ensure that the expected cost remains below a specified threshold. However, in the context of social navigation, few works have presented impressive results in complex scenarios with CRL. 
From our perspective, directly constraining collision rates, as seen in previous work \cite{zhou2023safe}, is not an optimal approach to formulate the problem. This approach makes the optimization problem difficult to solve since costs are only received at the end of episodes, similar to the challenges in sparse reward conditions. 
RL agents may find it difficult to understand what leads to collisions since the complexity is potentially high.
In our work, we propose to use uncertainty quantification obtained by ACI to design a buffer around pedestrians and constrain the cumulative intrusions into this buffer over an episode. 
Compared to previous methods that directly constrain collision rates, our method provides more behavior-level guidance to RL agents, offering rich cost feedback. 
We call our method \textit{spatial relaxation} in the context of social navigation. Our goal remains to improve safety, but by converting the constraints on collision rates to cumulative intrusions, the learning process becomes easier to converge without compromising safety. 
In our paper, we demonstrate that spatial relaxation achieves much better performance compared to directly imposing constraints on collision rates.

Integrating these two techniques, we introduce SoNIC (Safe \textbf{So}cial \textbf{N}avigation with Adaptive Conformal \textbf{I}nference and \textbf{C}onstrained RL), which can generate safe trajectories with minimal intrusions into pedestrians' paths, as illustrated in Fig. \ref{fig:teaser}. The main contributions of this paper are as follows:
\begin{itemize}
\item We develop a novel safe social navigation framework that integrates nonconformity scores generated by ACI with CRL, which not only augments the observations of RL agents but also directly guides their learning process and regulates their behaviors.
\item We propose a technique to increase the applicability of CRL in the context of social navigation by introducing spatial relaxation. Compared to previous methods, spatial relaxation provides richer cost feedback and facilitates convergence without sacrificing safety.
\item Our method achieves state-of-the-art (SOTA) performance on the CrowdNav benchmark in terms of both safety and adherence to social norms. Specifically, it obtains an 11.67\% higher success rate, 4.5 times fewer collisions, and 2.8 times fewer intrusions into human trajectories compared to the previous SOTA RL baseline. Moreover, our method demonstrates stronger robustness under two out-of-distribution (OOD) conditions.
\item Besides our extensive simulation experiments and ablation studies, we also develop a ROS2-based navigation system and deploy our methods on real robots with Mecanum kinematics. We find that, with basic clipping and smoothing, the policy network trained in simulation can be directly deployed on the robot, which exhibits polite and safe behaviors when interacting with both sparse and dense crowds.
\end{itemize}
\noindent
The main focus of this paper lies in addressing the safety challenges of navigating dense crowds through uncertainty-aware prediction and CRL and can adapt to conditions where static obstacles can be represented by a collection of static agents, as is a common practice of previous work \cite{chen2017decentralized}.
\section{Related Work}
\subsection{Social Robot Navigation}
Social robots are expected to interact with humans and complete various tasks, such as providing assistance \cite{francis2023principles}. Social navigation forms the foundation for accomplishing most high-level tasks. Robots are required to navigate in crowds, where the challenge of modeling dynamic human behavior makes navigation particularly difficult. It is crucial to capture the subtleties of human behavior, such as human intentions and interactions between agents \cite{li2020evolvegraph,girase2021loki,zhou2022grouptron,lange2023scene,li2023pedestrian}, and properly utilize them for effective robotic decisions.
Deep reinforcement learning (DRL) offers a potentially viable solution for solving the challenging navigation task \cite{li2023game,li2024interactive,li2024multi}.
Previous works on RL-based methods for social robots include capturing agent-agent interactions \cite{chen2019crowd,li2024multi} and the intentions of human agents \cite{liu2023intention}, incorporating these as predictions into RL policy networks. Our work takes a step further by quantifying the uncertainties of these predictions and guiding robot behavior based on the uncertainty quantification.

\subsection{Planning Under Uncertainty}
Trajectory planning under uncertainty has become a field that attracts growing attention. In optimization-based and search-based methods, researchers have attempted to integrate uncertainty quantification from perception and prediction into various controllers \cite{yang2023safe, dixit2023adaptive}. Both methods share the characteristic of easily adding constraints or shielding, allowing for explicit management of uncertainties. In contrast, our work focuses on enhancing the performance of RL agents by incorporating prediction uncertainties, as RL demonstrates stronger generalization ability compared to traditional methods, offering faster computation time and better performance in complex real-life scenarios that require long-horizon decision-making capabilities of agents. However, the organic integration between uncertainty quantification and RL can be challenging.
For instance, previous work has augmented observations of RL agents by inputting uncertainties to policy networks \cite{huang2023conformal} and post-editing actions generated by RL policies according to uncertainties to generate safe behaviors \cite{strawn2023conformal}. 
However, most of these works circumvent the direct guidance and regularization in the RL learning processes, resulting in RL agents that cannot fully leverage the uncertainty quantification results. More recently, Golchoubian et al. \cite{golchoubian2024uncertainty} combine prediction uncertainties into RL policies for low-speed autonomous vehicles and design a reward function to encourage the ego agent to avoid the uncertainty-augmented prediction area. However, their uncertainty metrics lack guaranteed coverage, and reward shaping can not always effectively regulate the behaviors of agents as expected \cite{sutton2018reinforcement}. Additionally, their methods were tested in a relatively simple environment with sparse human distribution.
In contrast, we use DtACI \cite{gibbs2024conformal} for uncertainty quantification, providing provable coverage and robustness against distribution shifts. Besides, we use CRL with spatial relaxation to effectively guide policy learning and validate our results in a complex environment with dense crowds.

\subsection{Safe Reinforcement Learning}
Safe RL enables incorporating safety constraints into DRL methods, allowing dangerous conditions to be avoided \cite{ji2023omnisafe}. Common techniques include state augmentation \cite{sootla2022saute}, adding safety layers or components to modify actions generated by unsafe RL agents \cite{yu2022towards}, and Lagrangian methods. Among these, Lagrangian methods are relatively easy to implement, can be applied to almost any RL algorithm, and outperform some more complex methods in benchmark tests \cite{ray2019benchmarking}.
In trajectory planning, previous work has validated that safe RL can fulfill safety constraints in some simple settings, such as simulation environments with sparse or static obstacles\cite{ray2019benchmarking, selim2022safe}. However, in these settings where environments are relatively easy, safe RL algorithms do not show significant performance advantages compared to vanilla RL algorithms. In contrast, we validate our method in complex environments for social navigation with dense moving pedestrians, and it outperforms previous SOTA results by a large margin.

\section{Preliminaries}
\subsection{Adaptive Conformal Inference}
Conformal methods can augment model predictions with a prediction set that is guaranteed to contain true values with a predefined coverage, enabling the quantification of uncertainties in a model-agnostic manner \cite{gibbs2024conformal}. Traditional split conformal prediction requires a calibration set and places high demands on the exchangeability between the test sample and the calibration samples. 
In contrast, adaptive conformal inference (ACI) can dynamically adjust its parameters to maintain coverage in an online and distribution-free manner \cite{gibbs2021adaptive}, making it appealing for time-sequential applications. Dynamically-tuned adaptive conformal inference (DtACI) \cite{gibbs2024conformal} further boosts the applicability and performance of ACI by running multiple prediction error estimators with different learning rates simultaneously. DtACI adaptively selects the best output based on its historical performance, eliminating the need to pre-acquire underlying data dynamics to achieve satisfying coverage.

\begin{figure*}[!tbp]
	\centering
	\includegraphics[width=1.0\textwidth]{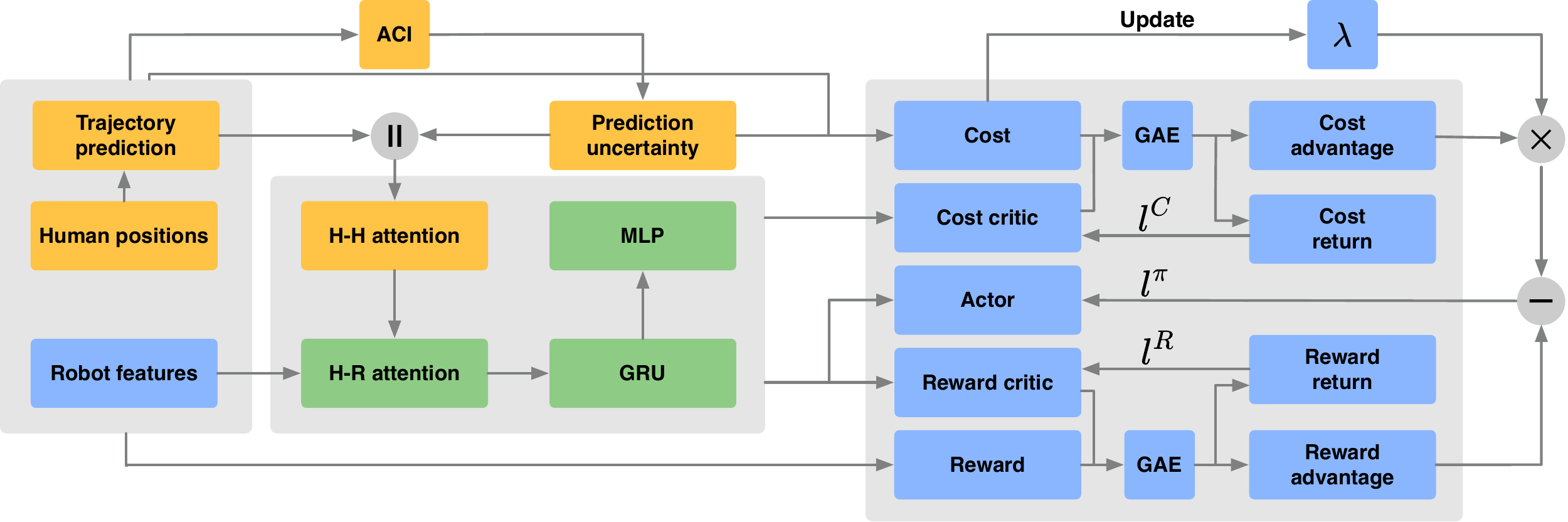}
	\caption{The overall pipeline of SoNIC. We mark components related to humans in yellow, components related to physical information and decision-making of the robot in blue, and fused features in green. We use ACI to quantify the prediction uncertainty of human trajectories and concatenate these metrics with predictions before inputting them into networks. The networks contain attention mechanisms for interactions between humans (H-H attention) and between humans and the ego robot (H-R attention). Prediction uncertainty combined with physical information is used for designing costs. For the CRL agent using PPO Lagrangian, the actor and reward critic share some layers while the cost critic uses a separate network. We adopt reward value loss $l^R$, action loss $l^{\pi}$, and cost value loss $l^C$ for updating the agent. More details about the architectures and training strategy can be found in Section IV.}
	\label{fig:overall_diagram}
\end{figure*}

\subsection{Constrained Reinforcement Learning}
CRL extends RL algorithms by incorporating constraints on the agent behavior. Unlike traditional Markov Decision Process (MDP) settings where agents learn behaviors only to maximize rewards, CRL is often formulated as a Constrained Markov Decision Process (CMDP). At time step \( t \), an agent chooses an action \( A_t \) under state \( S_t \), receives a reward \( R_t \), and incurs a cost \( C_t \), after which the environment transitions to the next state \( S_{t+1} \). In a CMDP, the objective is not only to find an optimal policy that maximizes rewards but also to manage costs associated with certain actions or states, which may be defined as quantities related to safety in the context of social navigation. This is generally represented as \cite{ray2019benchmarking}:
\begin{equation}
    \pi^*=\arg \max _{\pi \in \Pi^C} J^R(\pi),
\end{equation}
where \( J^R(\pi) \) is a reward-based objective function, and \( \Pi^C \) is the feasible set of policies that satisfy the constraints added to the problem. The goal of CRL is to ensure that costs remain within pre-defined thresholds while maximizing reward.

\section{Method}
\subsection{Problem Formulation}
The major task of social navigation is enabling robots to navigate through crowds and addressing the challenges of making appropriate decisions for successful navigation. In our setting, we have $H$ humans in the environment, each indexed by $h$, within an episode of horizon $T$. 
At each time step $t$, the ground-truth positions of humans are represented as $\mathbf{p}_{h}(t)$. We predict $K$ future steps (a larger $K$ means more extended future predictions) for each human's future trajectories to better understand their intentions. The prediction point for the $k$-th prediction of the $h$-th human is denoted as $\mathbf{p}_{h, k}(t)$, where $1 \leq h \leq H$ and $1 \leq k \leq K$.

We formulate the task as a CMDP, where the CRL agent is provided with observations of the state \(S_t\) at each timestep, which consists of two main parts. The first part includes physical information: the current positions of humans and the robot, and other quantities about the robot's dynamics. The second part comprises post-processed features generated by models, such as the predicted human trajectories. We denote physical state components of ego information as \(\mathbf{e}\), physical components of human information as \(\mathbf{h}\), and components generated by models as \(\mathbf{m}\). The complete state \(S_t\) is written as \(S_t = [\mathbf{e}, \mathbf{h}, \mathbf{m}]\).
The agent then generates action \(A_t = (v_x, v_y)\) to control the moving speed of the robot according to \(S_t\). This means we adopt holonomic kinematics, which aligns with previous work for easier comparisons of agent performance \cite{chen2019crowd, liu2023intention}. One common concern about holonomic kinematics is that it may not be easily transferred to real robots; however, we find that direct post-editing, including clipping and smoothing according to real-robot kinematics, of the actions produced by the agent can enable good sim-to-real performance in our experiments. For more details, please refer to our Sec.~\ref{sec:real_experiments}.
After taking an action, the environment transitions to the next state \(S_{t+1}\) based on the motion dynamics of humans and the robot. It then provides a reward \(R_t\) and a cost \(C_t\). We aim to obtain an optimal policy \(\pi(A_t \mid S_t)\) that maximizes rewards while satisfying cost constraints, ensuring that the expected cumulative cost over an episode is below a threshold.

\subsection{Method Overview}
The overall pipeline of SoNIC is illustrated in Fig.~\ref{fig:overall_diagram}. First, we employ two different trajectory predictors: the constant velocity predictor (CV) \cite{scholler2020constant} and the Gumbel social transformer (GST) predictor \cite{huang2021learning}. This is to demonstrate that our algorithm can adapt to both rule-based and learning-based predictors. Next, we use ACI to quantify prediction uncertainties and incorporate them into \(\mathbf{m}\); subsequently, we adopt a policy network with a combined attention mechanism as proposed in \cite{liu2023intention} to process these uncertainties along with other features. Among the ACI methods, DtACI is the most advanced, providing state-of-the-art performance in adapting to distribution shifts online \cite{gibbs2024conformal}. Therefore, we choose DtACI for uncertainty quantification to further enhance SoNIC's adaptability. Finally, we employ CRL to guide the agents' behavior using the uncertainty quantification results. Instead of directly applying constraints to the collision rate, we impose constraints on cumulative intrusions of the robot into other agents’ spatial buffers. This approach, referred to as spatial relaxation in our paper, provides behavior-level guidance and effectively addresses the issue of sparse constraint feedback, thereby improving upon previous methods that constrained result-oriented metrics such as the collision rate.

\subsection{Rule-Based and Learning-Based Trajectory Prediction}
Trajectory prediction of other agents serves as an explicit representation of their intentions, thereby enabling the decision model to act more effectively \cite{liu2023intention}. In practice, different prediction models emphasize various aspects, such as inference speed or prediction accuracy, and researchers may select prediction models according to their specific needs. Our system adapts seamlessly to different prediction models, including both learning-based and rule-based trajectory predictors, and can mitigate the adverse effects of incorrect predictions on the subsequent decision-making process.

For the rule-based prediction model, we use the CV model to obtain simple and effective estimates of future human states by extrapolating the current states of human agents based on their velocities. This can be represented by the following equation:
\begin{equation}
    \mathbf{p}_{h, k}(t) = \mathbf{p}_{h}(t) +  \mathbf{v}_{h}(t) \times k \Delta t,
\end{equation}
where $\mathbf{p}_{h, k}(t)$ denotes the $k$-th predicted position of the $h$-th human, $ \mathbf{p}_{h}(t)$ is the current position of the human, $ \mathbf{v}_{h}(t)$ is their current velocity, $\Delta t$ is the time step length, $1 \leq h \leq H$, and $1 \leq k \leq K$.

For learning-based prediction models, given a crowd of $H$ humans, each indexed by $h\in\{1,\ldots,H\}$, we denote by $\mathbf{p}_h(t)$ the ground-truth 2D position of the $h$-th human at time step $t$. The objective is to model the distribution of their $K$-step future trajectories according to their $K^{\prime}$-step histories. In other words, the prediction models need to learn the probability:
\begin{equation} \resizebox{0.91\hsize}{!}{ $
    P\Big(\big\{\mathbf{p}_{h}(t+1),\ldots,\mathbf{p}_{h}(t+K)\big\}_{h=1}^{H} \;\Big|\;\Big\{\mathbf{p}_{h}(t-K^{\prime}+1),\ldots,\mathbf{p}_{h}(t)\Big\}_{h=1}^{H}\Big),
    $}
\end{equation}
where $P$ stands for probability, $K$ denotes the number of future steps to predict, and $K^{\prime}$ means the number of past steps to use for prediction. Our overall goal is to make the predictions generated by the model, $\mathbf{p}_{h, k}(t)$, be better aligned with ground truth future locations.

We choose the GST predictor \cite{huang2021learning} as the learning-based prediction model, which is designed to address the challenges of partially detected pedestrians and redundant interaction modeling. GST dynamically generates sparse interaction graphs, ensuring that the target agent focuses only on the most relevant neighbors, thereby reducing the influence of redundant information. Its architecture is adapted for the efficient encoding of pedestrian features. GST demonstrates adaptability and robustness in high-density crowds.

\subsection{ACI for Quantifying Prediction Uncertainty}
After obtaining the $K$-step future predictions, we quantify the prediction uncertainty using DtACI~\cite{gibbs2024conformal}, an adaptive conformal inference algorithm that adapts effectively to distribution shifts and thus serves as a good fit for online uncertainty estimation in social navigation.

For each prediction step of each pedestrian, we run $M$ prediction error estimators simultaneously. At time step $t$, we calculate the actual prediction error $\delta_{h, k}$ between the current position and the predicted position made at time step $t-k$ for the $k$-th prediction step of the $h$-th human:
\begin{equation} 
    \delta_{h, k}(t) = \| \mathbf{p}_{h}(t) - \mathbf{p}_{h, k}(t-k) \|_2,
\end{equation}
where $\mathbf{p}_{h}(t)$ is the actual position of the $h$-th human at time $t$, $\mathbf{p}_{h, k}(t-k)$ is the $k$-step future predicted position of the $h$-th human made at time $t-k$, and $\delta_{h, k}(t)$ is calculated as the L2 norm of the difference between the two values.

We then update the estimated prediction error generated by the $m$-th estimator for the $h$-th human according to:
\begin{equation} 
    \hat{\delta}^{(m)}_{h, k}(t) = \hat{\delta}^{(m)}_{h, k}(t-1) 
    \;-\; \gamma^{(m)} \bigl(\alpha - \operatorname{err}_{h, k}^{(m)}(t)\bigr),
\end{equation}
where $\hat{\delta}^{(m)}_{h, k}$ represents the estimated prediction error of the $m$-th estimator corresponding to a $k$-step ahead prediction for the $h$-th human, $\gamma^{(m)}$ is the learning rate of the $m$-th prediction error estimator for all humans and predictions, $\alpha$ is the coverage parameter, and 
\begin{equation}
    \operatorname{err}^{(m)}_{h, k}(t) := 
    \begin{cases} 
        1, & \text{if } \hat{\delta}^{(m)}_{h, k}(t-1) < \delta_{h, k}(t), \\ 
        0, & \text{if } \hat{\delta}^{(m)}_{h, k}(t-1) \geq \delta_{h, k}(t).
    \end{cases}
\end{equation}
Intuitively, when $\hat{\delta}^{(m)}_{h, k} < \delta_{h, k}$, implying that the estimation of the $m$-th estimator does not cover the actual prediction error, $\hat{\delta}^{(m)}_{h, k}$ will increase by $(1-\alpha) \gamma^{(m)}$; otherwise, $\hat{\delta}^{(m)}_{h, k}$ will decrease by $\alpha \gamma^{(m)}$. Therefore, the updating speed differs in these two cases. According to quantile regression \cite{gibbs2024conformal}, $\hat{\delta}^{(m)}_{h, k}$ will converge to be no less than $(1-\alpha)$ of all actual prediction errors, thereby achieving $(1-\alpha)$ coverage.

Since we run $M$ prediction error estimators with different learning rates simultaneously, for each error estimator, after taking in the actual prediction error $\delta_{h, k}$ and updating the estimated prediction error for the next step, we evaluate the errors of each estimator and update the probability distribution for choosing the next output prediction error estimator by
\begin{equation} \resizebox{0.91\hsize}{!}{ $
    w_{h, k}^{(m)} \leftarrow (1 - \sigma)\,\frac{w_{h, k}^{(m)} 
    \exp\bigl(-\eta\, \ell(\delta_{h, k}, \delta_{h, k}^{(m)})\bigr)}
    {\sum_{j=1}^{M} w_{h, k}^{(j)} 
    \exp\bigl(-\eta\,\ell(\delta_{h, k}, \delta_{h, k}^{(j)})\bigr)} 
    \;+\; \frac{\sigma}{M},
    $ }
\end{equation}
\begin{equation}
    p_{h, k}^{(m)} \leftarrow \frac{w_{h, k}^{(m)}}
    {\sum_{j=1}^{M} w_{h, k}^{(j)}},
\end{equation}
where we ignore the explicit time-step notation $t$ for simplicity, and use the arrow to indicate the replacement of values. The weight $w_{h, k}^{(m)}$ corresponds to the probability $p_{h, k}^{(m)}$ for the $m$-th prediction error estimator, $\sigma$ and $\eta$ are hyperparameters of DtACI for adjusting the speed at which weights change. $\ell(\delta_{h, k}, \delta_{h, k}^{(m)})$ is the loss function used to measure the estimation error, where we adopt the pinball loss:
\begin{equation} \resizebox{0.91\hsize}{!}{$
    \ell(\delta_{h, k}, \delta_{h, k}^{(m)}) =
    \begin{cases}
        \alpha \,\bigl(\delta_{h, k} - \delta_{h, k}^{(m)}\bigr), 
            & \text{if } \delta_{h, k} \ge \delta_{h, k}^{(m)},\\
        (\alpha - 1)\,\bigl(\delta_{h, k} - \delta_{h, k}^{(m)}\bigr), 
            & \text{if } \delta_{h, k} < \delta_{h, k}^{(m)}.
    \end{cases} $}
\end{equation}
Each time we estimate the prediction uncertainty, we treat
$\hat{\delta}_{h, k}$ as a discrete random variable taking values in the set
$\{\hat{\delta}_{h, k}^{(1)}, \ldots, \hat{\delta}_{h, k}^{(M)}\}$.
The probability mass function of $\hat{\delta}_{h, k}$ is:
\begin{equation} \resizebox{0.91\hsize}{!}{$
    P\bigl(\hat{\delta}_{h, k} = \hat{\delta}_{h, k}^{(m)}\bigr) 
    \;=\; p_{h, k}^{(m)},
    \quad m = 1,\ldots,M, \quad \sum_{m=1}^{M} p_{h, k}^{(m)} \;=\; 1.
    $}
\end{equation}
In other words,
\(\hat{\delta}_{h, k}\) follows a categorical distribution whose probability mass function is
\(\bigl\{p_{h,k}^{(1)}, \dots, p_{h,k}^{(M)}\bigr\}\).

\subsection{Policy Network Structure}
Once we have obtained the trajectory prediction results and the corresponding prediction uncertainty, we concatenate the uncertainty quantification with the predicted trajectory before feeding it into the attention layers~\cite{liu2023intention}. This allows the RL agents to account for the prediction uncertainty in their decision-making process, as shown in Fig.~\ref{fig:overall_diagram}. The first block is the human-human attention (H-H attention in Fig.~\ref{fig:overall_diagram}), which models each human as a separate node to capture interactions among humans. Next, we fuse the robot features (including velocity, heading, positions, and goal) into the attention blocks through human-robot attention (H-R attention) to obtain fused feature embeddings capturing the interactions between humans and the ego robot. We then process these fused embeddings together with the robot features and concatenate them to form the GRU input, thereby capturing temporal information. Lastly, the final fused features are passed to the actor and critic networks for further processing. For more details about the policy network structure, please refer to~\cite{liu2023intention}.

\subsection{CRL with Spatial Relaxation}
Deploying CRL to social navigation by directly constraining safety metrics like collision rates may lead to sparse feedback conditions, as costs are generated only at the end of episodes. This results in similar issues to the classic sparse reward problem, making it difficult for agents to learn optimal state and action values due to the potentially high complexity from actions to outcomes. 
To address this issue, we propose to provide agents with dense behavior-related guidance through costs. In the context of social navigation, we constrain the cumulative intrusions into a buffer zone based on the prediction uncertainty quantified by ACI around pedestrians and set constraints on these intrusions. We call our method \textit{spatial relaxation} as it allows for a more flexible approach to managing safety. Instead of directly limiting collision rates, our method tolerates minor intrusions within a controlled buffer zone, making the optimization problem easier to solve while maintaining a high level of safety.

We design the buffer of pedestrians as a combination of a circular area around the human's current position and an ACI-generated area around $K^\prime$ ($K^\prime \leq K$) steps of predictions. Since we have $H$ human agents in the environment, the two parts of the buffer are defined as follows:
\begin{equation}
\resizebox{0.85\hsize}{!}{$
D_i(\mathbf{p}_{\text{ego}}) = \left\{ \mathbf{p}_{\text{ego}} : \left| \mathbf{p}_{\text{ego}} - \mathbf{p} \right| \leq r_i \right\}, \; \mathbf{p} \in P_i, \; i = 1, 2,
$}
\end{equation}
\begin{equation}
    P_1 = \{ \mathbf{p}_h \}, \; P_2 = \{ \mathbf{p}_{h, k} \}, \; 1 \leq h \leq H, \; 1 \leq k \leq K^\prime,
\end{equation}
\begin{equation}
    r_1 = r_{\text{ego}} + r_h + r_{\text{buffer}}, \quad r_2 = r_{\text{ego}} + r_h + \hat{\delta}_{h, k},
\end{equation}
where $D_1$ is the subarea that considers buffers around the current positions of humans and $D_2$ is the subarea that considers buffers around the predicted positions of humans. If the current center position of the ego robot $\mathbf{p}_{\text{ego}}$ is in either $D_1$ or $D_2$, an intrusion occurs. For the computation, we consider the distance between the center positions of agents and prediction points. For the buffers corresponding to the current positions of humans, $r_1$ and $r_2$ are the corresponding distance thresholds for $D_1$ and $D_2$. $r_{\text{ego}}$ is the radius of the ego robot, $r_h$ is the radius of the $h$-th human, $r_{\text{buffer}}$ is the radius of the buffer around the current positions of humans, and $\hat{\delta}_{h, k}$ is the prediction uncertainty generated by DtACI for the $k$-th prediction point of the $h$-th human.

At each time step $t$, we iterate through all buffers of all humans and calculate the maximum intrusion, denoted as $d_{\text{intru}, t}$. For an episode with a horizon of $T$, we have
\begin{equation} \label{eq:optimization_goal}
    \max_{\pi} \sum_{t=0}^{T} R_t(S_t, A_t) \quad \text{s.t.} \quad \sum_{t=0}^{T} d_{\text{intru}, t} = \tilde{d},
\end{equation}
where $\tilde{d}$ is a pre-defined threshold. We formulate the cost term $C_t$ using the intrusions into $D = D_1 \cup D_2$:
\begin{equation}
    C_t(S_t, A_t) = \mu d_{\text{intru}, t},
\end{equation}
where $\mu$ is a constant. Our reward includes three components:
\begin{equation}
R_t\left(S_t, A_t\right) = 
\begin{cases} 
R_{\text{success}}, & \text{if } p_{\text{ego}} \in S_{\text{goal}}, \\ 
R_{\text{collision}}, & \text{if } p_{\text{ego}} \in S_{\text{fail}}, \\ 
R_{\text{potential}}, & \text{otherwise},
\end{cases}
\end{equation}
where $S_{\text{goal}}$ means the robot reaches the goal, $S_{\text{fail}}$ means the robot collides with other pedestrians, and $R_{\text{potential}}$ provides a dense reward that drives the ego robot to approach the goal, proportional to the distance the ego robot approaches the goal compared to the previous time step \cite{liu2023intention}. 

Please note that we do not claim to fully replace conventional reward shaping with CRL. Instead, we retain the essential components of the reward function to allow the robot to learn basic social navigation principles while using CRL to adaptively handle safety-related behavior. Specifically, we separate out the portion of the reward function concerning behavior patterns and treat it as a cost signal, which is then adaptively optimized via the mechanism introduced by CRL.

According to Lagrange duality \cite{ji2023omnisafe}, instead of directly solving the optimization goal in Eq. (\ref{eq:optimization_goal}), we maximize a derived \textit{Lagrangian function}:
\begin{equation}
\begin{aligned}
 L(\pi, \lambda) &= \sum_{t=0}^{T} R_t(S_t, A_t) - \lambda \left( \sum_{t=0}^{T} d_{\text{intru}, t} - \tilde{d} \right) \\
 &= \sum_{t=0}^{T} \left[ R_t(S_t, A_t) - \lambda d_{\text{intru}, t} + \lambda \frac{\tilde{d}}{T} \right],
 \label{eq:lagrangian_function}
\end{aligned}
\end{equation}
where $\lambda$ is the Lagrangian multiplier. Then, the optimization problem can be viewed as an unconstrained  RL problem that aims to maximize a combined reward in Eq. (\ref{eq:lagrangian_function}).

In our work, we use the PPO Lagrangian \cite{ray2019benchmarking} for maximizing the Lagrangian function. We set up two critics to compute the state value for reward and the state value for cost. The loss functions for the two critics are defined as
\begin{align}
    l^R_t &= c_1 (V^R_{\theta_1}(S_t) - V_t^{\text{targ}, R})^2, \\
    l^C_t &= c_2 (V^C_{\theta_2}(S_t) - V_t^{\text{targ}, C})^2,
\end{align}
where $c_1$ and $c_2$ are constants, $V^R_{\theta_1}(S_t)$ and $V^C_{\theta_2}(S_t)$ are network-generated value estimates for reward and cost, respectively, and $V_t^{\text{targ}, R}$ and $V_t^{\text{targ}, C}$ are the corresponding target values for temporal difference updates.

As for the policy network, the action loss is similar to the form in PPO \cite{schulman2017proximal} where we employ the combined advantage:
\begin{equation}
    \hat{A}^{\prime}_t = \frac{\hat{A}^R_t - \lambda \hat{A}^C_t}{1+\lambda},
\end{equation}
which can be interpreted as a linear combination of advantages, corresponding to the Lagrangian function in Eq.~\ref{eq:lagrangian_function}, scaled by $(1 + \lambda)$ to stabilize training and aid convergence. Notably, the constant term \(\lambda \tilde{d}/T\) in the Lagrangian does not contribute to the advantage function, as it is independent of the policy's actions and thus its advantage is always zero. The action loss function can then be written as
\begin{equation}
\resizebox{0.85\hsize}{!}{$
l_t^{\pi} = \hat{\mathbb{E}}_t \left[\min \left(r_t(\theta_3) \hat{A}^{\prime}_t, \operatorname{clip}\left(r_t(\theta_3), 1 - \epsilon, 1 + \epsilon\right) \hat{A}^{\prime}_t\right)\right],
$}
\end{equation}
where $r_t(\theta_3)$ represents the change ratio between the updated and old policy, $\epsilon$ is the predefined clip ratio, and $\hat{A}^{\prime}_t$ represents the estimated value advantage function using Generalized Advantage Estimation (GAE) \cite{schulman2015high} at time step $t$. In our implementation, the parameters of the actor and reward critic, $\theta_1$ and $\theta_3$, share some network structures, as shown in Fig. \ref{fig:overall_diagram}, but the parameters of the cost critic are independent. We set the distribution entropy of the stochastic action to be constant and do not adopt the entropy loss from the original PPO. We found that this setting works better for SoNIC when training with large batches.

Lastly, we update the Lagrangian multiplier \(\lambda\) using gradient descent so that the averaged cumulative intrusion converges to the predefined value \(\tilde{d}\). The loss function \cite{ji2023omnisafe} for updating \(\lambda\) is defined as
\begin{equation}
l_t^{\lambda} = -\lambda(\bar{C} - \tilde{d}_C),
\end{equation}
where \(\bar{C}\) is the mean episode cost, which in practice is calculated by averaging over the past few episodes, and \(\tilde{d}_C\) is the cost limit and is proportional to $\tilde{d}$. Intuitively, when \(\bar{C}\) is greater than \(\tilde{d}_C\) (i.e., intrusions are frequent), \(\lambda\) will increase according to gradient descent, leading the RL agents to consider cost advantages more when updating the policy and thus, actions with larger costs will be less preferred, and vice versa. Note that we aim to train RL agents with satisfactorily safe behaviors; Therefore, in practice, we set \(\tilde{d}_C\) to be small.

\section{Experiments}
\label{sec:experiments}
\subsection{Simulation Settings and Evaluation Metrics}
\label{sec:simulation_settings}
We train and evaluate our RL agents using CrowdNav \cite{chen2019crowd}, a simulator with flexible settings that can simulate complex pedestrian behaviors. Our training environment involves 20 humans and a robot in a \SI{12}{m} $\times$ \SI{12}{m} area with randomized positions and goals. The distance between the robot's start position and its goal is greater than \SI{8}{m} but smaller than \SI{12}{m}. The robot radius is set to \SI{0.2}{m}, while the human radius is randomly sampled between \SI{0.3}{m} and \SI{0.5}{m}. The robot's maximum speed is set to \SI{1.0}{m/s}. For every episode, we set the time limit to be \SI{50}{s}.

Some features of CrowdNav that make the navigation task in CrowdNav challenging include:
\begin{itemize}
    \item \textit{High density of humans in the scenarios:} As mentioned above, we set 20 humans in a rather space-limited scenario, the overall scenarios are highly crowded.
    \item \textit{Sudden changes in humans' moving directions and velocities:} The goals of the humans may change suddenly with a probability of 50\% every five time steps, which subsequentially changes the moving directions and velocities of the humans, which makes the movement of humans harder to predict and poses challenges to both prediction models and decision models.
    \item \textit{Robot is invisible to humans:} In this setting, the robot is assumed to be invisible to humans, meaning that humans do not actively avoid it. This design ensures that the robot does not exploit human avoidance behaviors, thereby minimizing unnecessary interventions in human behavior. For comparison, we have also trained SoNIC models in a setting where humans interact with the robot, and the results are presented in Appendix \ref{sec:visible_robot}. These results demonstrate that models trained in the invisible robot setting can naturally adapt to the visible robot setting, whereas the reverse is not true.
\end{itemize}

For our training and test settings described in both in-distribution and OOD test settings in Sec. \ref{sec:in-dist} to Sec. \ref{sec:ood}, we keep \textit{all} of the above settings identical. However, we set the following settings differently:
\begin{itemize}
    \item \textit{Training and in-distribution test settings.} For in-distribution settings, meaning that the test settings are the same as the training process, we set the human behavior policy to be Optimal Reciprocal Collision Avoidance (ORCA) \cite{van2008reciprocal} and the maximum speed of humans is randomly sampled between \SI{0.5}{m/s} and \SI{1.5}{m/s}.
    \item \textit{OOD scenarios mixed with 20\% rushing humans.} For this condition, we also set the human policy to be ORCA \cite{van2008reciprocal}. However, we randomly select 20\% of the human agents to have a maximum speed of \SI{2.0}{m/s}, while the remaining agents have their speeds randomly sampled between \SI{0.5}{m/s} and \SI{1.5}{m/s}. This setup is designed to test whether our method can handle more challenging scenarios similar to real-life conditions, where some individuals may hurry through the environment.
    \item \textit{OOD scenarios with SF pedestrian model.} In this setting, we change the pedestrian behavior policy to be the Social Force (SF) \cite{helbing1995social} model, which is to test whether our method can generalize to scenarios with different human behavior patterns.
\end{itemize}

We adopt the standard evaluation metrics \cite{liu2023intention} including Success Rate (SR), Collision Rate (CR), Timeout Rate (TR), Navigation Time (NT), Path Length (PL), Intrusion Time Ratio (ITR), and Social Distance (SD). The detailed definitions can be found in Appendix \ref{sec:metrics}. For all of these metrics, SR and TR directly evaluate the safety performance of the robot. TR, NT, and PL measure the efficiency of the policy-generated paths, and ITR and SD measure the politeness of the paths. We believe that we should put SR and TR as the most important metrics to be compared. Policies with lower SR or higher CR, even if they might perform better on other metrics, are still not as good as those with higher SR and lower CR.

\subsection{Baselines and Ablation Models}

\begin{table*}[!tbp]
    \setlength{\tabcolsep}{1mm}
    \centering
    \caption{In-distribution test results}
    \fontsize{5.5}{6}\selectfont
    \resizebox{1.0\textwidth}{!}{
        \renewcommand{\arraystretch}{0.6}
       \begin{tabular}{
            m{1.9cm}<{\raggedright} | m{1.3cm}<{\centering} m{1.3cm}<{\centering} m{1.3cm}<{\centering} m{1.0cm}<{\centering} m{1.0cm}<{\centering} m{1.3cm}<{\centering} m{1.0cm}<{\centering} m{1.3cm}<{\centering}
        }
            \toprule
            Methods & \textbf{SR}$\uparrow$ & \textbf{CR}$\downarrow$ & \textbf{TR}$\downarrow$ & \textbf{NT}$\downarrow$ & \textbf{PL}$\downarrow$ & \textbf{ITR}$\downarrow$ & \textbf{SD}$\uparrow$ \\
            \midrule 
            SF \cite{helbing1995social} & 15.60\% & 21.44\% & 62.96\% & 30.23 & 34.64 & 3.78\% & 0.42 \\
            ORCA \cite{van2008reciprocal} & 67.84\% & 27.52\% & 4.64\% & 22.80 & 19.74 & \textbf{1.10}\% & \textbf{0.50} \\
            CrowdNav++ \cite{liu2023intention} & 86.80\% & 13.20\% & \textbf{0.00}\% & 14.05 & 19.81 & 7.71\% & 0.42 \\
            \midrule                                       
            RL (w/o ACI) & 92.67$\pm$1.51\% & 7.33$\pm$1.51\% & \textbf{0.00$\pm$0.00}\% & \textbf{12.89$\pm$0.10} & \textbf{19.42$\pm$0.12} & 10.62$\pm$0.57\% & 0.39$\pm$0.01 \\
            RL (w/ ACI) & 94.08$\pm$1.18\% & 5.92$\pm$1.18\% & \textbf{0.00$\pm$0.00}\% & 13.35$\pm$0.26 & 19.88$\pm$0.25 & 8.81$\pm$0.67\% & 0.40$\pm$0.00 \\

            RL (w/ ACI, Penalty) &    
            94.91$\pm$0.61\% & 
            5.09$\pm$0.61\% & 
            \textbf{0.00$\pm$0.00}\% & 
            13.47$\pm$0.40& 
            20.02$\pm$0.44& 
            6.68$\pm$0.46\% & 
            0.42$\pm$0.01\\
            
            CRL (w/ ACI, on CR) & 93.79$\pm$1.57\% & 5.71$\pm$1.00\% & 0.51$\pm$0.68\% & 19.94$\pm$1.17 & 26.32$\pm$1.32 & 4.74$\pm$0.72\% & 0.40$\pm$0.00 \\
            \midrule
            SoNIC (w/ CV) & 96.03$\pm$1.14\% & 3.73$\pm$1.24\% & 0.24$\pm$0.24\% & 17.88$\pm$0.60 & 24.51$\pm$0.76 & 2.40$\pm$0.22\% & 0.45$\pm$0.00 \\
            SoNIC (w/ GST) & \textbf{96.93$\pm$0.68}\% & \textbf{2.93$\pm$0.61}\% & 0.13$\pm$0.12\% & 17.54$\pm$0.86 & 24.27$\pm$0.85 & 2.72$\pm$0.16\% & 0.44$\pm$0.00 \\
            \bottomrule
        \end{tabular}
    }
    \label{tab:in_distribution_result}
\end{table*}

\begin{table*}[!tbp]
    \setlength{\tabcolsep}{1mm}
    \centering
    \caption{OOD test results - mixed with 20\% rushing humans}
    \fontsize{5.5}{6}\selectfont
    \resizebox{1.0\textwidth}{!}{
        \renewcommand{\arraystretch}{0.6}
       \begin{tabular}{
            m{1.9cm}<{\raggedright} | m{1.3cm}<{\centering} m{1.3cm}<{\centering} m{1.3cm}<{\centering} m{1.0cm}<{\centering} m{1.0cm}<{\centering} m{1.3cm}<{\centering} m{1.0cm}<{\centering} m{1.3cm}<{\centering}
        }
            \toprule
            Methods & \textbf{SR}$\uparrow$ & \textbf{CR}$\downarrow$ & \textbf{TR}$\downarrow$ & \textbf{NT}$\downarrow$ & \textbf{PL}$\downarrow$ & \textbf{ITR}$\downarrow$ & \textbf{SD}$\uparrow$ \\
            \midrule 
            SF \cite{helbing1995social} & 12.24\% & 19.12\% & 68.64\% & 32.06 & 36.15 & 5.31\% & 0.40 \\
            ORCA \cite{van2008reciprocal} & 60.32\% & 34.96\% & 4.72\% & 23.41 & 19.84 & \textbf{2.95} \% & \textbf{0.48} \\
            CrowdNav++ \cite{liu2023intention} & 71.92 \% & 28.08\% & \textbf{0.00} \% & 14.11 & 18.87 & 13.72\% & 0.39 \\
            \midrule                                       
            RL (w/o ACI) & 74.19$\pm$1.26\% & 25.81$\pm$1.26\% & \textbf{0.00$\pm$0.00}\% & \textbf{13.31$\pm$0.30} & \textbf{18.20$\pm$0.10} & 18.23$\pm$0.31\% & 0.37$\pm$0.00 \\
            RL (w/ ACI) & 76.96$\pm$4.29\% & 23.04$\pm$4.29\% & \textbf{0.00$\pm$0.00}\% & 14.13$\pm$0.32 & 19.04$\pm$0.14 & 15.84$\pm$1.14\% & 0.37$\pm$0.01 \\
            RL (w/ ACI, Penalty) &    
            80.53$\pm$1.43\% & 
            19.47$\pm$1.43\% & 
            \textbf{0.00$\pm$0.00}\% & 
            14.27$\pm$0.37& 
            19.62$\pm$0.30& 
            12.13$\pm$0.26\% & 
            0.39$\pm$0.00\\
            CRL (w/ ACI, on CR) & 82.53$\pm$4.36\% & 17.31$\pm$4.31\% & 0.16$\pm$0.16\% & 19.99$\pm$1.30 & 25.10$\pm$1.94 & 8.94$\pm$1.77\% & 0.38$\pm$0.01 \\
            \midrule
            SoNIC (w/ CV) & 87.07$\pm$0.89\% & \textbf{12.75$\pm$1.21}\% & 0.19$\pm$0.32\% & 18.74$\pm$0.31 & 24.57$\pm$0.66 & 5.29$\pm$0.20\% & 0.40$\pm$0.00 \\
            SoNIC (w/ GST) & \textbf{87.17$\pm$4.14}\% & \textbf{12.75$\pm$4.00}\% & 0.08$\pm$0.14\% & 18.32$\pm$1.01 & 24.04$\pm$0.85 & 6.82$\pm$1.36\% & 0.38$\pm$0.00 \\
            \bottomrule
        \end{tabular}
    }
    \label{tab:ood_rushing_result}
\end{table*}

\begin{table*}[!tbp]
    \setlength{\tabcolsep}{1mm}
    \centering
    \caption{OOD test results - SF pedestrian model}
    \fontsize{5.5}{6}\selectfont
    \resizebox{1.0\textwidth}{!}{
        \renewcommand{\arraystretch}{0.6}
       \begin{tabular}{
            m{1.9cm}<{\raggedright} | m{1.3cm}<{\centering} m{1.3cm}<{\centering} m{1.3cm}<{\centering} m{1.0cm}<{\centering} m{1.0cm}<{\centering} m{1.3cm}<{\centering} m{1.0cm}<{\centering} m{1.3cm}<{\centering}
        }
            \toprule
            Methods & \textbf{SR}$\uparrow$ & \textbf{CR}$\downarrow$ & \textbf{TR}$\downarrow$ & \textbf{NT}$\downarrow$ & \textbf{PL}$\downarrow$ & \textbf{ITR}$\downarrow$ & \textbf{SD}$\uparrow$ \\
            \midrule 
            SF \cite{helbing1995social} & 12.08\% & 6.72\% & 81.20\% & 29.76 & 40.56 & 1.60\% & 0.45 \\
            ORCA \cite{van2008reciprocal} & 92.56\% & 4.88\% & 2.56\% & 22.36 & 21.91 & \textbf{0.72}\% & \textbf{0.48} \\
            CrowdNav++ \cite{liu2023intention} & 92.48\% & 7.52\% & \textbf{0.00} \% & 13.24 & 19.47 & 7.31\% & 0.41 \\
            \midrule                                       
            RL (w/o ACI) & 95.68$\pm$0.89\% & 4.32$\pm$0.89\% & \textbf{0.00$\pm$0.00}\% & \textbf{12.35$\pm$0.21} & \textbf{18.99$\pm$0.16} & 9.98$\pm$0.49\% & 0.39$\pm$0.00 \\
            RL (w/ ACI) & 97.41$\pm$0.81\% & 2.59$\pm$0.81\% & \textbf{0.00$\pm$0.00}\% & 13.08$\pm$0.24 & 19.81$\pm$0.24 & 8.07$\pm$0.43\% & 0.40$\pm$0.01 \\
            RL (w/ ACI, Penalty) &    
            98.40$\pm$0.56\% & 
            1.60$\pm$0.56\% & 
            \textbf{0.00$\pm$0.00}\% & 
            13.10$\pm$0.48& 
            19.88$\pm$0.48& 
            6.32$\pm$0.05\% & 
            0.41$\pm$0.01
            \\
            CRL (w/ ACI, on CR) & 96.29$\pm$1.62\% & 2.99$\pm$1.00\% & 0.72$\pm$0.63\% & 20.39$\pm$1.50 & 27.20$\pm$1.67 & 4.17$\pm$0.92\% & 0.40$\pm$0.00 \\
            \midrule
            SoNIC (w/ CV) & 98.48$\pm$0.92\% & 1.39$\pm$0.82\% & 0.13$\pm$0.12\% & 19.02$\pm$0.45 & 25.84$\pm$0.54 & 2.04$\pm$0.25\% & 0.43$\pm$0.01 \\
            SoNIC (w/ GST) & \textbf{98.96$\pm$0.52}\% & \textbf{1.04$\pm$0.52}\% & \textbf{0.00$\pm$0.00}\% & 18.18$\pm$0.84 & 25.05$\pm$0.79 & 2.66$\pm$0.45\% & 0.42$\pm$0.00 \\
            \bottomrule
        \end{tabular}
    }
    \label{tab:ood_sf_result}
\end{table*}

Our baselines include ORCA \cite{van2008reciprocal}, SF \cite{helbing1995social}, and CrowdNav++ \cite{liu2023intention}. ORCA and SF are classic algorithms in obstacle avoidance, while CrowdNav++ represents the previous SOTA algorithm in social navigation. For CrowdNav++, we directly use their pre-trained model for testing.

To validate the effectiveness of ACI and CRL with spatial relaxation, our ablation settings include:
\begin{itemize}
    \item \textit{RL (w/o ACI):} This setting is similar to CrowdNav++ but with substantial hyperparameter changes, such as a simpler reward function and constant action noise.
    \item \textit{RL (w/ ACI):} In this variant, we augment the observations (as used in RL (w/o ACI)) with human prediction uncertainty generated by ACI.
    \item \textit{RL (w/ ACI, Penalty):} Similar to RL (w/ ACI), but with an additional reward penalty proportional to the intrusions into the buffers around both the current human positions and their predictions (the same penalty per step as costs used in SoNIC (w/ CV) and SoNIC (w/ GST)).
    \item \textit{CRL (w/ ACI, on CR):} In this setting, we also augment observations with ACI-generated uncertainty but employ CRL by imposing constraints on the collision rate.
    \item \textit{SoNIC (w/ CV):} This variant employs both ACI and CRL with spatial relaxation, using a simple CV prediction model for predicting human future positions.
    \item \textit{SoNIC (w/ GST):} Similar to SoNIC (w/ CV), but using the GST predictor \cite{huang2021learning} for human prediction.
\end{itemize}

\subsection{Implementation Details}
Under each ablation setting, we train our models using three different random seeds on an NVIDIA RTX 4090 GPU, keeping all other hyperparameters consistent except those we intend to compare. Key common settings include: 
\begin{itemize}
    \item All models use human predictions as part of the input observations and employ a pre-trained GST predictor with fixed parameters to generate five steps of human trajectory predictions, except for SoNIC (w/ CV).
    \item For the training parameters, we leverage the parallel training feature of PPO by setting the batch size to 32 and running 128 parallel environments. The learning rate is set to $3\times10^{-5}$, and the clip parameter is set to 0.08 for both the actor and the reward critic. The learning rate for the cost critic is set to $1.5\times10^{-5}$. We run $2\times10^7$ training steps to obtain the final policy model for testing and real-robot deployment. In our training, varying entropy is removed and the action noise, which is normally distributed, is only added during training and removed during testing and deployment.
    \item We set the buffers around humans, \(r_{\text{buffer}}\), to \SI{0.25}{m} and calculate costs based on the intrusions into the buffer and the first two prediction uncertainty areas, while inputting predictions for all five future steps into the policy network with a coverage parameter \(\alpha = 0.1\), corresponding to 90\% coverage in prediction errors.
    \item For CRL (w/ ACI, on CR), the cost limit is set to correspond to a 3\% collision rate. For SoNIC (w/ CV) and SoNIC (w/ GST), the cost limit corresponds to cumulative intrusions of approximately \SI{0.16}{m} per episode on average (a cumulative value), which translates to a cost limit \(\tilde{d}_C\) of 0.4. The initial value of the Lagrangian multiplier is set to 0.1, and its learning rate is $1.6 \times 10^{-3}$.
    \item For DtACI hyperparameters, the initial prediction errors are set to \SI{0.1}{m}, \SI{0.2}{m}, \SI{0.3}{m}, \SI{0.4}{m}, and \SI{0.5}{m} for 1- to 5-step-ahead predictions, respectively. We employ three error estimators with learning rates \(\gamma\) of 0.05, 0.1, and 0.2 for each DtACI estimator.
    \item For the specific reward and cost functions, we set the success reward $R_{\text{success}}$ to $+10$, the collision penalty $R_{\text{collision}}$ to $-20$, and the dense potential reward that guides the robot's movement to $2 \Delta d_{\text{forward}}$, where \(\Delta d_{\text{forward}}\) measures the distance the robot moves toward the goal compared to the previous step. The cost \(C_t(S_t, A_t)\) is defined as $2.5  d_{\text{intru}, t}$, where \(d_{\text{intru}, t}\) represents the maximum intrusion into the buffers of the current human positions and the first two prediction uncertainty areas.
\end{itemize}

\begin{figure*}[!tbp]
    \centering
    \includegraphics[width=1.0\textwidth]{figures/QualitativeAnalysis.pdf}
    \caption{Visualization of test results for different cases. Pedestrians are shown in blue, the robot in yellow, and the goal is represented by the orange star. The spatial buffers based on uncertainty quantification are depicted as light blue circles around humans, while the subareas considered in CRL are a slightly deeper shade. (a) SoNIC (w/ GST) performing in an in-distribution environment, successfully navigating to the goal. (b) CrowdNav++ performing in the same episode but failing to complete the task. In this subfigure, the light blue circles indicate prediction lines rather than spatial buffers. (c) SoNIC (w/ GST) performing in an OOD environment with rushing humans. (d) SoNIC (w/ GST) performing in an OOD environment with the SF pedestrian model.}
    \label{fig:QualitativeAnalysis}
\end{figure*}

\subsection{In-distribution Test Results}
\label{sec:in-dist}
\textit{1) Quantitative Analysis:} For all the test results shown in Tables \ref{tab:in_distribution_result}-\ref{tab:ood_sf_result}, we evaluate 1250 samples across 5 random test seeds and calculate the mean performance and standard deviations of models trained with 3 different training seeds. 
The test results under the same setting as the training environment (i.e., in distribution) are shown in Table \ref{tab:in_distribution_result}. 

From these results, we can see that rule-based methods SF and ORCA are not capable enough of handling complex scenarios in our settings, as indicated by their high CR. However, both methods have low ITR and high SD since they are specifically designed to avoid intrusions. 
When comparing all the RL-based methods, both SoNIC (w/ GST) and SoNIC (w/ CV) outperform other methods in safety metrics and ITR, indicating that SoNIC can generate both safe and polite trajectories that cause minimal intrusions to pedestrians. Although SoNIC has shorter TR, NT, and PL compared to RL models without constraints, this tradeoff is reasonable considering the overall improvement, especially since SR and CR are the most critical metrics. Notably, SoNIC (w/ GST) achieves an 11.67\% increase in success rate, decreases collisions by 4.51 times, and reduces intrusions by 2.83 times compared to CrowdNav++.
While SoNIC (w/ GST) shows marginally better performance than SoNIC (w/ CV) in safety metrics, the small gap suggests that our methods effectively mitigate prediction errors and adapt well to simple prediction models.

Interestingly, we observe a performance improvement in RL (w/ ACI) compared to RL (w/o ACI). This suggests that informing the agent about prediction uncertainties can enhance its ability to handle complex situations, though not as effectively as SoNIC. Furthermore, the performance increase in RL (w/ ACI, Penalty) compared to RL (w/ ACI) demonstrates that appropriate reward shaping can improve the performance of RL agents to some extent, but the improvement remains smaller than that of SoNIC (w/ GST).

Additionally, CRL (w/ ACI, on CR) does not outperform RL (w/ ACI), suggesting that direct constraints on collision rates are ineffective in improving safety. 
SoNIC surpasses CRL (w/ ACI, on CR) across all metrics, which strongly supports the effectiveness of the spatial relaxation proposed in this paper.

\textit{2) Qualitative Analysis:} We visualize the behaviors of SoNIC (w/ GST) and CrowdNav++ in the same episode in Fig. \ref{fig:QualitativeAnalysis}(a) and Fig. \ref{fig:QualitativeAnalysis}(b), respectively. At the beginning, CrowdNav++ decides to approach the goal directly. However, as the pedestrians move, they gradually surround the robot, leaving it with no escape route, which leads to an almost inevitable collision for CrowdNav++. 
In contrast, SoNIC chooses to move the robot out of the crowds to prevent potential collisions or intrusions into the spatial buffers of pedestrians from the start. 
From step 24, we can see that SoNIC rapidly reacts to a human with a sudden change in direction, and the expanding uncertainty area due to prediction errors accumulated in the past few steps helps the robot perform an avoidance maneuver. 
Similar conditions occur again at steps 61 to 63, where the robot successfully avoids a human who changes direction before finally moving to the goal. This demonstrates that SoNIC is not only adept at making decisions that are beneficial in the long term but also capable of rapid execution in emergency situations.

\subsection{OOD Test Results}
\label{sec:ood}
\textit{1) OOD Scenarios Mixed with 20\% Rushing Humans:} 
In this setting, we set 20\% of the human agents to have a maximum speed of \SI{2.0}{m/s}, which is 33\% higher than the maximum speed that may appear during the training phase. 

From the results in Table \ref{tab:ood_rushing_result}, we observe that all methods exhibit degraded performance compared to in-distribution conditions. 
SF and ORCA show lower success rates, but the overall performance drop in RL-based methods is more pronounced. This indicates that rule-based methods maintain baseline performance but fail to excel in complex scenarios. 
CrowdNav++, RL (w/o ACI), RL (w/ ACI), and RL (w/ ACI, Penalty) experience significant performance declines, with SR drops of 14.88\%, 18.48\%, 17.12\%, and 14.38\%, respectively. 
In contrast, CRL (w/ ACI, on CR), SoNIC (w/ CV), and SoNIC (w/ GST) exhibit smaller SR drops of 11.26\%, 8.96\%, and 9.76\%, respectively. The smaller performance degradation in CRL methods suggests their potential to mitigate challenges arising from OOD conditions. 

Notably, SoNIC (w/ CV) and SoNIC (w/ GST) still achieve the best results, outperforming other RL algorithms by an even larger margin. Furthermore, the performance gap between SoNIC (w/ CV) and SoNIC (w/ GST) narrows from 0.90\% to 0.1\% in SR and CR compared to standard in-distribution conditions. SoNIC (w/ CV) demonstrates higher stability across random seeds, which is likely due to the CV predictor's robustness in OOD conditions.

We visualize a scenario performed by SoNIC (w/ GST) in Fig. \ref{fig:QualitativeAnalysis}(c), where two rushing humans are moving toward the agent at step 32. Due to the GST predictor facing the challenges of OOD conditions, the uncertainty area generated by ACI becomes much larger than in in-distribution cases, which demonstrates the effectiveness of using DtACI in adapting to sudden distributional shifts of pedestrian behaviors. The robot then chooses to navigate between these two agents through the gap between the uncertainty areas at step 34 and successfully escapes to a safe area at step 38.
\begin{figure}[!tbp]
	\centering
	\includegraphics[width=1.0\columnwidth]{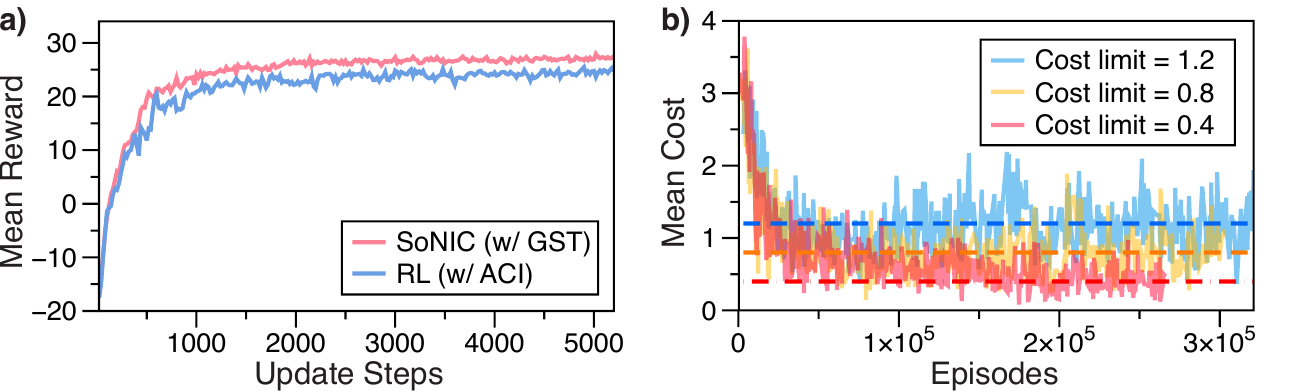}
	\caption{Convergence analysis of SoNIC. (a) The learning curves of SoNIC (w/ GST) and RL (w/ACI). SoNIC (w/ GST) shows faster convergence with higher rewards. (b) The cost curves of SoNIC (w/ GST) with different cost limits. The average costs across episodes can approximately approach the predefined cost limits, which are shown by the dashed lines.
    }
	\label{fig:curves}
\end{figure}

\textit{2) OOD Scenarios with SF Pedestrian Model:} 
In this setting, we change the behavior policy of all human agents from ORCA to SF. From the results in Table \ref{tab:ood_sf_result}, we can see that all methods perform better than in in-distribution conditions, which indicates that this setting is easier than the ORCA setting.
Although SF still suffers from a low SR, ORCA becomes even better than CrowdNav++ in SR, CR, ITR, and SD, showing that the performance advantage of CrowdNav++ depends on conditions. 
In this case, CRL (w/ ACI, on CR) performs worse than RL (w/ ACI) again, which indicates that policies learned by directly constraining CR may be too conservative to grasp opportunities to fulfill the tasks. 
The two SoNIC methods achieve almost perfect results in terms of SR, CR, and TR, implying that SoNIC \textit{adapts well} to OOD scenarios caused by different behavior models.

We visualize a scenario performed by SoNIC (w/ GST) in Fig. \ref{fig:QualitativeAnalysis}(d). When three humans approach each other, SoNIC maintains a conservative distance from the crowds and successfully escapes afterward.

\subsection{Convergence Analysis}
We present the learning curves of SoNIC (w/ GST) in Fig. \ref{fig:curves}(a), alongside RL (w/ ACI) for comparison. The results indicate that SoNIC (w/ GST) demonstrates a smoother learning process and achieves higher rewards. Furthermore, as shown in Fig. \ref{fig:curves}(b), the average episodic costs of SoNIC (w/ GST) converge to different cost limit values. This provides an effective mechanism for adjusting the aggressiveness of robot policies, as further discussed in Appendix \ref{sec:aggressiveness}.

\begin{figure*}[!tbp]
	\centering
	\includegraphics[width=1.0\textwidth]{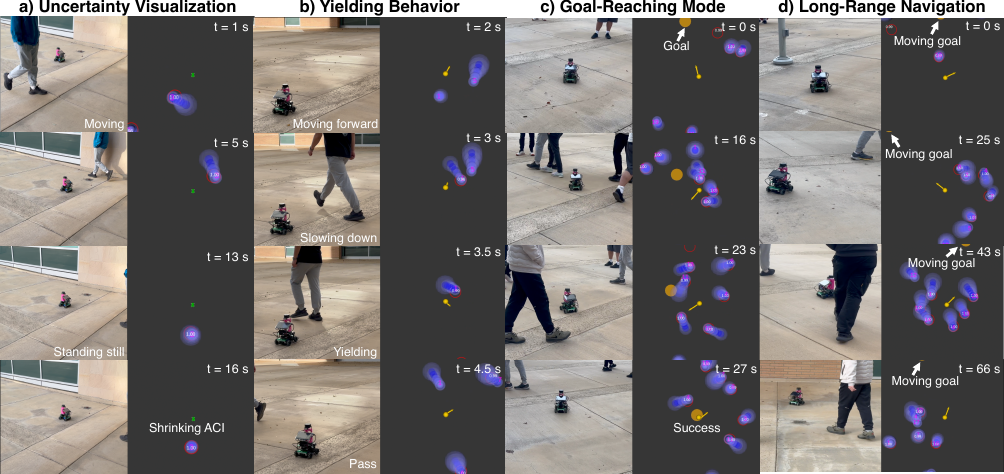}
	\caption{We deploy our methods on a ROSMASTER X3 with Mecanum wheels using the ROS2 system. For the four subplots, the left sides display photos taken from the experiments, and the right sides show visualizations in RViz. In the RViz visualizations, the red circles represent detection results, and the white numbers inside the red circles indicate the output probabilities of the detection model. The purple numbers correspond to the indices generated by the tracker. The prediction lines are shown in blue, and the prediction uncertainties are depicted in semi-transparent light blue. In (a), the green robot represents the robot’s current location. In (b)-(d) where the decision node is enabled, the yellow sphere indicates the robot’s position and the yellow arrow represents the command output from the decider node. The orange circle indicates the goal position. (a) In the uncertainty visualization, the human initially walks around the robot and then stands still behind it. The uncertainty area adjusts dynamically based on the prediction accuracy. (b) The robot equipped with SoNIC demonstrates stable yielding behavior when interacting with humans. (c) In goal-reaching mode, the robot navigates through crowds and successfully reaches its goal. (d) In long-range navigation mode, the moving goal consistently guides the robot’s movement.
    }
	\label{fig:experiments}
\end{figure*}

\subsection{Real Robot Experiments} \label{sec:real_experiments}
We deploy our approach on a ROSMASTER X3 robot equipped with Mecanum wheels, enabling flexible movement and independent control of $v_x$ and $v_y$. 
The robot connects to a laptop with an Nvidia RTX 3070 GPU (mobile version) via a router. 
The SoNIC models are trained in CrowdNav simulations with holonomic dynamics and are deployed directly to the robot system \textit{without} further fine-tuning. 

The robot uses a 2D LiDAR (RPLIDAR-A1) for human detection via a learning-based detector. The LiDAR operates at a scanning frequency of approximately \SI{6}{Hz}, which also limits the frequencies of the tracking, prediction, and decision making modules. Despite this, our experiments demonstrate that the robot makes effective navigation decisions in crowded environments, indicating that the system generalizes well to cost-efficient robots. Human detection is done with a pre-trained DR-SPAAM model \cite{jia2020dr}, tracking is done with SORT \cite{bewley2016simple}, prediction is done with the GST model \cite{huang2021learning}, and decision making is powered by SoNIC (w/GST). For further details about our system, please refer to Appendix \ref{sec:ros2}, and computational speed analyses for the core functions of each module are provided in Appendix \ref{sec:compute_speed}.

We conduct the experiments on a large outdoor terrace covering an area greater than \SI{15}{m} $\times$ \SI{20}{m}. The robot operates in two movement modes:
\begin{itemize}
    \item \textit{Goal-reaching mode:} The robot navigates to a predetermined target point, which is the original setting in the standard CrowdNav training phase.
    \item \textit{Long-range navigation mode:} The goal of the robot is dynamically updated based on its movement. This approach addresses the limitation that, during training, the goal inputs to the policy networks are constrained to a limited range. As a result, distant goals beyond \SI{12}{m} may negatively impact policy performance. By introducing a continuously updated moving target, we show that the robot can reliably navigate distances exceeding \SI{20}{m} through experiments. This method has the potential to be extended to even longer distances if the system is fully integrated onboard in the future.
\end{itemize}

We show several representative real-world testing scenarios and results in Fig. \ref{fig:experiments}.
Specifically, we first visualize the prediction uncertainties using RViz in Fig. \ref{fig:experiments}(a). Before $t= \SI{13}{s}$, the human walks around the robot, and the robot successfully detects the human's position while visualizing the predicted trajectory and the associated uncertainty area. 
Since the prediction model does not fully capture the human's circular walking pattern, the long-horizon prediction uncertainty area is significantly larger than that of short-horizon predictions. After $t= \SI{13}{s}$, the human stands still behind the robot. Initially, the prediction uncertainty area is large because the uncertainty results are carried over from iterative processes. Gradually, the uncertainty area shrinks to reflect more precise prediction uncertainties. After approximately 3 seconds, the uncertainty area stabilizes and reduces to almost zero. This process demonstrates that by using DtACI, the system can effectively adapt to different conditions and dynamically reflect prediction uncertainties. This enables safer decision making without being overly conservative.

In Fig. \ref{fig:experiments}(b), by allowing the robot to interact with a single agent, we observe that the robot can generate stable yielding behaviors. At $t= \SI{2}{s}$, the robot detects that a human is coming along a path that will intersect with its original moving trajectory. It begins by slowing down to reduce the risk of a collision ($t= \SI{3}{s}$). At $t= \SI{3.5}{s}$, the robot adjusts its direction, moving towards the backside area of the human to avoid intrusions. At $t= \SI{4.5}{s}$, the robot successfully passes.

In Fig. \ref{fig:experiments}(c), the robot navigates in goal-reaching mode while interacting with humans along its path. We observe that the robot encounters very dense crowds at $t = \SI{16}{s}$ and $t = \SI{23}{s}$. Throughout this process, the robot makes appropriate decisions to minimize intrusions into humans’ trajectories while progressing toward its goal, and it ultimately reaches the goal despite the dense human interactions.

In Fig. \ref{fig:experiments}(d), we implement the long-range navigation mode, enabling the robot to navigate distances beyond those encountered during training. In this mode, the moving goal is dynamically updated to remain \SI{5}{m} ahead of the robot’s position along the longitudinal axis, while its lateral position remains fixed. Under these conditions, the robot moves forward while actively avoiding dynamic obstacles, as demonstrated at $t = \SI{25}{s}$. Although the robot temporarily deviates laterally to avoid collisions, the fixed lateral component of the moving goal serves as a stable reference, helping the robot stay within a reasonable lateral range.

Overall, our system demonstrates robust decision-making capabilities in crowded environments. By quantifying prediction uncertainty and incorporating it into buffer representations rather than relying solely on prediction lines, the robot develops more adaptive spatial awareness and exhibits consistent, safe, and socially aware behaviors.

\section{Limitations and Future Work}
\label{sec:limitations}
Our current work focuses on social navigation in open, human-populated spaces with highly dynamic interactions. Extending our method to handle static obstacles with arbitrary shapes, like walls and furniture, would require integrating additional perception and mapping modules, such as incorporating SLAM into our system and using multi-modal data as input to the policy network. 
While this is beyond the scope of the current work, it represents a clear direction for future research.
Another limitation stems from the perception module. Although SoNIC’s decision making is robust, failure cases sometimes arise due to perception errors such as human miss-detection and the 2D LiDAR’s sensitivity to varying lighting conditions. 
To address these issues, we plan to incorporate cameras into our perception system, leverage sensor fusion techniques, and upgrade the current 2D LiDAR to a model with a higher spinning frequency and improved robustness, thereby providing sufficient safety redundancy.

Finally, we believe that the long-range navigation mode presented in our paper offers interesting insights into how to flexibly adjust policy inputs for more adaptable deployment in navigation scenarios. We plan to further explore this aspect and integrate it with long-range planning modules, testing its practical value in large-scale, real-world environments. 

\section{CONCLUSION}
In this paper, we present SoNIC, a novel framework for safe social navigation that integrates ACI with CRL to generate safe robot trajectories that adhere to social norms. Our framework not only enhances the observations of RL agents and but also guides their learning process and regulates their behaviors. We propose the spatial relaxation technique to increase the applicability of CRL in social navigation, which can provide rich feedback and facilitate convergence without sacrificing safety. 
Our method achieves SOTA performance in both safety and adherence to social norms on the CrowdNav benchmark and shows strong robustness to OOD scenarios. Finally, we develop a real-robot system that spans from perception to decision making, and deploy it on a real robot equipped with Mecanum kinematics. In our experiments, the robot demonstrates safe and socially compliant navigation behaviors when interacting with human crowds.

\bibliographystyle{plainnat}
\bibliography{references}
\clearpage

\appendix
\subsection{Evaluation Metrics}
\label{sec:metrics}
The evaluation metrics used in our paper are introduced in detail as follows:
\begin{itemize}
    \item \textit{Success Rate (SR):} 
    SR measures the ratio of the number of successful episodes (i.e., the robot reaching the goal within the time limit of \SI{50}{s})
    to the total number of test episodes, i.e., 
    \( \text{SR} = N_{\text{success}} / N_{\text{total}} \).

    \item \textit{Collision Rate (CR:}
    CR measures the ratio of episodes with at least one collision
    to the total number of test episodes, i.e.,
    \( \text{CR} = N_{\text{collision}} / N_{\text{total}} \).

    \item \textit{Timeout Rate (TR):}
    TR measures the ratio of episodes that fail to reach the goal before the time limit
    to the total number of test episodes, i.e.,
    \( \text{TR} = N_{\text{timeout}} / N_{\text{total}} \).

    \item \textit{Navigation Time (NT):}
    NT is the average time required for the robot to reach the goal (computed only over \textit{successful} episodes), i.e.,
    \( \text{NT} = \bigl(1 / N_{\text{success}}\bigr) \sum_{k=1}^{N_{\text{success}}} T_k \),
    where \(T_k\) is the time the robot took to reach the goal in the \(k\)-th successful episode.

    \item \textit{Path Length (PL):}
    PL is the total distance traveled by the robot in an episode, accumulated step by step.
    We report the average path length across all episodes (including collisions and timeouts), i.e.,
    \( \text{PL} = \bigl(1 / N_{\text{total}}\bigr) \sum_{i=1}^{N_{\text{total}}} \text{dist}_i \),
    where \(\text{dist}_i\) is the total distance traveled in episode \(i\).

    \item \textit{Intrusion Time Ratio (ITR):}
    ITR is the fraction of time steps in an episode
    where the robot intrudes into the ground-truth future positions of humans (\textit{Danger} event triggered).
    Its final value is an average over all episodes:
    \begin{equation} \resizebox{0.80\hsize}{!}{ $
        \text{ITR} 
        = \frac{1}{N_{\text{total}}} \sum_{i=1}^{N_{\text{total}}}
        \Bigl(\frac{\#\,\text{danger steps in episode } i}{\text{total steps in episode } i} 
        \times 100\%\Bigr).
    $}
    \end{equation}

    \item \textit{Social Distance (SD):} SD is the average of the minimal distances between the robot and any human,
    computed only at the ``danger steps.'' We log the minimal distance whenever a \textit{Danger} event occurs, and then take the average over all episodes.
\end{itemize}
\subsection{ACI Effectiveness}
\begin{figure}[!tbp]
	\centering
	\includegraphics[width=1.0\columnwidth]{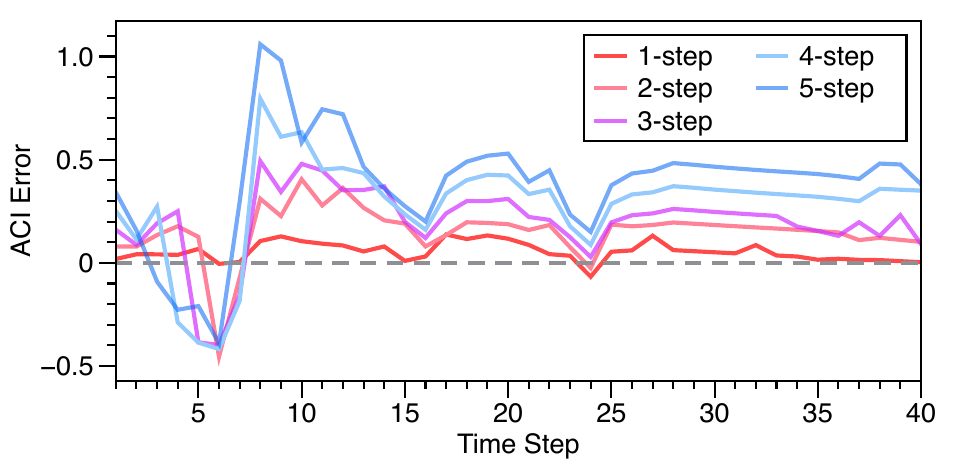}
	\caption{Visualization of ACI errors (i.e., estimated prediction error minus actual prediction error) for one pedestrian's five prediction steps during the time period it is within the observable area of the robot. ACI provides valid coverage when the ACI error is greater than 0.
    }
	\label{fig:ACIError}
\end{figure}

To validate the effectiveness of ACI in covering actual prediction errors, we visualize the ACI errors of one human in Fig. \ref{fig:ACIError}. At the beginning of this trajectory, the ACI errors for multi-step predictions are large because the GST predictor lacks sufficient information to accurately predict human positions, leading to large actual prediction errors that our initial ACI values do not cover properly. However, after several steps, ACI quickly adapts to the actual prediction error and achieves adequate coverage. Additionally, if the coverage remains sufficient, the ACI error will decrease to ensure that the uncertainty estimation is not too conservative.
\subsection{Tuning Aggressiveness of Trajectories with Cost Limits}
\label{sec:aggressiveness}
Although the main results of SoNIC presented in our paper focus on low cost limits to ensure safe navigation through crowds, we also observe interesting testing results with varying cost limits that allow for tuning the aggressiveness of trajectories, as shown in Table \ref{tab:tune_aggressiveness}. These results are obtained by testing across five seeds, with 250 test samples for each seed, using models trained with the same training seed. We find that as the cost limits increase, the trajectories become gradually more aggressive, reflected by an increase in ITR. This increased aggressiveness also improves efficiency, as indicated by reductions in NT and PL. However, it comes at the expense of key metrics such as SR and CR. Therefore, for a single best policy, we recommend keeping the cost limit as low as possible, provided the policy can still converge.

\begin{table}[!tbp]
    \setlength{\tabcolsep}{0.4mm}
    \centering
    \caption{Performance of SoNIC (w/ GST) Under Different Cost Limits}
    \fontsize{5.5}{6}\selectfont
    \resizebox{0.48\textwidth}{!}{
        \renewcommand{\arraystretch}{0.6}
       \begin{tabular}{
            m{0.8cm}<{\centering} | m{0.8cm}<{\centering} m{0.6cm}<{\centering} m{0.6cm}<{\centering} m{0.6cm}<{\centering} m{0.6cm}<{\centering} m{0.6cm}<{\centering} m{0.6cm}<{\centering} m{0.6cm}<{\centering}
        }
            \toprule
            Cost limit & \textbf{SR}$\uparrow$ & \textbf{CR}$\downarrow$ & \textbf{TR}$\downarrow$ & \textbf{NT}$\downarrow$ & \textbf{PL}$\downarrow$ & \textbf{ITR}$\downarrow$ & \textbf{SD}$\uparrow$ \\
            \midrule 
            0.4 & 96.96\% & 2.80\% & 0.24\% & 18.52 & 25.24 & 2.73\% & 0.43 \\
            0.6 & 96.72\% & 3.04\% & 0.24\% & 16.16 & 22.96 & 3.59\% & 0.44 \\
            0.8 & 94.56\% & 5.44\% & 0.00\% & 13.60 & 20.33 & 5.61\% & 0.43 \\
            1.0 & 94.32\% & 5.68\% & 0.00\% & 13.53 & 20.19 & 6.62\% & 0.42 \\
            1.2 & 93.76\% & 6.24\% & 0.00\% & 12.97 & 19.45 & 7.31\% & 0.41 \\
            \bottomrule
        \end{tabular}
    }
    \label{tab:tune_aggressiveness}
\end{table}

\subsection{ROS2 System for Real Robot Deployment}
\label{sec:ros2}
We develop a full ROS2 system from perception to decision making, including four main nodes:
\begin{itemize}
    \item \textit{Detector:} This node employs a pretrained DR-SPAAM model \cite{jia2020dr} to detect human agents from 2D LiDAR point clouds. The model leverages a lightweight neural network, enabling real-time detection on resource-constrained devices. According to the original paper, it achieves an accuracy metric of $AP_{0.5} = 70.3\%$.
    \item \textit{Tracker:} This node uses the SORT \cite{bewley2016simple} algorithm for tracking, which simply aims at assigning indices to detected agents. To generate consistent tracking results even when receiving noisy detections, we assign large values to the measurement noise covariance matrix \(\mathbf{R}\) and the process noise covariance matrix \(\mathbf{Q}\) of the underlying Kalman filter to accommodate detection jitter. We also set a higher noise coefficient for the position and scale components of the Kalman filter and a smaller value for the velocity components to allow more flexibility in position and scale updates. In the initial state covariance matrix \(\mathbf{P}\), the velocity components are given a large uncertainty, and the entire matrix is scaled accordingly so the filter can quickly adapt during early tracking. Besides, in the SORT tracker, we only keep trajectories whose ratio of valid detections over the entire detected period is above 0.9, filtering out spurious or short-lived detections. At each timestep, we run the Kalman filter's inference step to infer the positions of tracked agents even when they are not correctly detected in certain frames.
    \item \textit{Predictor:} This node uses the GST \cite{huang2021learning} model as the trajectory planner, integrating the DtACI module to directly apply uncertainty quantification after obtaining prediction lines for the pedestrians. The parameters and model weights are consistent with those used during training.
    \item \textit{Decider:} This node receives information from the upper-level nodes and determines the output action commands ($v_x$ and $v_y$ for velocities) for the controller. We utilize the controller integrated into the ROSMASTER X3 system by publishing \verb|\cmd_vel|. The decider supports three modes: goal-reaching mode, long-range navigation mode, and manual control mode. The selection of these modes is managed by another utility node, \textit{Command Listener}, which listens to user input and communicates the commands to the decider.
\end{itemize}
\subsection{Computational Speed Analysis}
\label{sec:compute_speed}
\begin{figure}[!tbp]
	\centering
	\includegraphics[width=1.0\columnwidth]{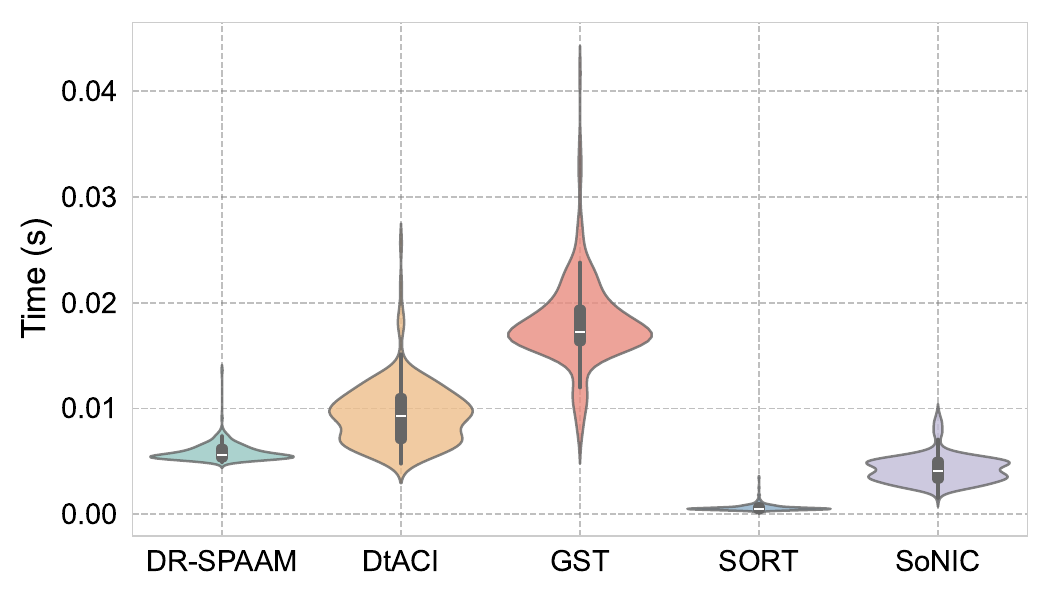}
	\caption{Visualizations of computational time distributions for five key modules of our ROS2 system. The data samples were collected during experiments in dense human environments. The shape of each violin plot represents the density of computation times, with wider areas indicating a higher concentration of values. The white dots at the center of each shape denote the mean computation time for the respective module.
    }
	\label{fig:computational_time}
\end{figure}

We log the time consumption of key functions across different nodes, including the inference times of the detection model (DR-SPAAM), prediction model (GST), and decision model (SoNIC), as well as the update times for the tracker (SORT) and uncertainty quantifier (DtACI) in dense crowd experiments, and visualize the results in Fig. \ref{fig:computational_time}. The visualization shows that GST has the longest average computation time and the largest variance. In contrast, DR-SPAAM and SORT maintain low and stable computation times, making them efficient for real-time applications. While DtACI has relatively high computational costs due to its pure Python implementation, its average time remains below \SI{0.01}{s}, well within acceptable limits. This overhead can be greatly reduced by transitioning to more efficient languages like C++. Despite this, the worst-case computation time for all main modules stays below 0.1 seconds, meeting and exceeding our system’s decision frequency requirements.

\subsection{Visible Robot Settings}
\label{sec:visible_robot}
We also train the SoNIC (w/ GST) model in the CrowdNav environments where the robot is visible to humans, and these humans actively avoid the robot. We cross-validate the results in both visible and invisible robot settings, as summarized in Tables~\ref{tab:visible_CR}--\ref{tab:visible_SR}. Specifically, we evaluate the models under five different testing seeds, each with 250 test samples, for three models trained under three distinct training seeds. From these results, we observe that although models trained in the visible robot setting achieve low collision rates in their in-distribution settings, they fail to generalize when humans do not react to the robot. In such scenarios, nearly half of the test cases result in collisions. In contrast, models trained in the invisible robot setting maintain low collision rates in both testing scenarios and, surprisingly, exhibit even lower collision rates when the robot is visible than those trained in the visible robot setting. In real-world situations, some pedestrians may ignore the robot (e.g., when they are using their mobile phones while walking). This mixture of reactive and non-reactive pedestrians poses significant challenges for models that assume all humans will actively avoid the robot. Thus, training in an environment where humans do not react to the robot proves to be more robust, as it better accounts for scenarios in which pedestrians may fail to notice or choose to ignore the robot.

\begin{table}[!tbp]
    \centering
    \caption{Cross-Validation of SoNIC (w/ GST) for Robot Visibility Settings Using Success Rates}
    \fontsize{5.5}{6}\selectfont
    \resizebox{0.40\textwidth}{!}{
    \renewcommand{\arraystretch}{1.0}
    \begin{tabular}{%
        m{1.2cm}<{\centering} | 
        m{1.2cm}<{\centering} 
        m{1.2cm}<{\centering}
    }
    \toprule
    \diagbox[width=1.2cm]{\textbf{Test}}{\textbf{Train}} 
        & \textbf{Invisible} 
        & \textbf{Visible} \\
    \midrule
    \textbf{Invisible} & \textbf{96.93$\pm$0.68}\% & 50.48$\pm$0.49\%\\
    \midrule
    \textbf{Visible}   &  \textbf{99.92$\pm$0.14}\%& 99.07$\pm$0.76\%\\
    \bottomrule
    \end{tabular}
    } 
    \label{tab:visible_CR}
\end{table}

\begin{table}[!tbp]
    \centering
    \caption{Cross-Validation of SoNIC (w/ GST) for Robot Visibility Settings Using Collision Rate}
    \fontsize{5.5}{6}\selectfont
    \resizebox{0.40\textwidth}{!}{
    \renewcommand{\arraystretch}{1.0} 
    \begin{tabular}{
        m{1.2cm}<{\centering} | 
        m{1.2cm}<{\centering} 
        m{1.2cm}<{\centering} 
    }
    \toprule
    \diagbox[width=1.2cm]{\textbf{Test}}{\textbf{Train}} 
        & \textbf{Invisible} 
        & \textbf{Visible} \\
    \midrule
    \textbf{Invisible} & \textbf{2.93$\pm$0.61}\% & 49.52$\pm$0.49\%\\
    \midrule
    \textbf{Visible}   &  \textbf{0.08$\pm$0.14}\%& 0.93$\pm$0.76\%\\
    \bottomrule
    \end{tabular}
    }
    \label{tab:visible_SR}
\end{table}
\end{document}